# Approximate Discrete Probability Distribution Representation using a Multi-Resolution Binary Tree

David Bellot and Pierre Bessière
Gravir/IMAG CNRS and INRIA Rhône-Alpes
Zirst - 655 avenue de l'Europe - Montbonnot
38334 Saint Ismier - France

**Abstract**

*Computing and storing probabilities is a hard problem as soon as one has to deal with complex distributions over multiples random variables. The problem of efficient representation of probability distributions is central in term of computational efficiency in the field of probabilistic reasoning. The main problem arises when dealing with joint probability distributions over a set of random variables: they are always represented using huge probability arrays. In this paper, a new method based on a binary-tree representation is introduced in order to store efficiently very large joint distributions. Our approach approximates any multi-dimensional joint distributions using an adaptive discretization of the space. We make the assumption that the lower is the probability mass of a particular region of feature space, the larger is the discretization step. This assumption leads to a very optimized representation in term of time and memory. The other advantages of our approach are the ability to refine dynamically the distribution every time it is needed leading to a more accurate representation of the probability distribution and to an anytime representation of the distribution.*

## 1. Introduction

Computing and storing probabilities is a hard problem as soon as one has to deal with complex distributions over multiples random variables. In the field of Bayesian networks [12], this problems arises frequently. Indeed, a Bayesian network stores a joint probability distribution over a set of random variables as a product of simple conditional probabilities such as $P(X_1 \ldots X_n) = \prod_{i=1}^{n} P(X_i|pa(X_i))$ [14]. The term $P(X_i|pa(X_i))$ denotes the probability of $X_i$ given the parents of $X_i$ in the network (a graph). The parents of a random variable are those which have a direct influence on $X_i$ [11]. Beyond the problem of inference and learning with Bayesian networks and more general probabilistic graphical models [7], the problem of efficient representation of probability distributions is also central in term of computational efficiency of the legacy algorithms to deal with such probabilistic models. The main problem arises when dealing with joint probability distributions over a set of random variables. Indeed, many inference and learning algorithms in the domain of Bayesian reasoning decompose the calculus over a set of small subset of random variables. The size of those subsets depends on the topology of the graphical model and inference and learning algorithms are more efficient with small subsets [5, 8]. However, the space and time needed to compute with those subsets depends also on the dimension of the variables. Then the problem is to have an efficient representation of small joint probability distributions, i.e. of multi-dimensional probability distributions.

We will restrict our purposes to discrete and discretized continuous random variables. In this case, the joint distributions are usually represented using probability tables (or conditional probability tables in the case of Bayesian networks) as defined in [6]. A probability table is simply an array of probability values which has a size equal to the number states of the variable (or the product of states of each variables in the case of a joint distribution). Besides, it is easy to compute with discrete variables but the complexity of the calculus grows exponentially with the number of variables. For example, if $A, B$ and $C$ are three discrete random variables with ten states (say $a_1, \ldots, a_{10}, b_1, \ldots, b_{10}$ and $c_1, \ldots, c_{10}$), then the probability table used to represent $P(ABC)$ stores $10^3$ values.

Another example is the conditional tables used to represent, for example, $P(A|B_1 \ldots B_{31})$. Its associated probability table has the same space complexity as $P(A, B_1 \ldots B_{31})$. If $A, B_1 \ldots B_{31}$ are binary random variables, $2^{32}$ values are needed to store the table. This is huge and clearly outperforms the capabilities of actual computers.

A more efficient way to store those probability tables is needed in order to deal with such very large joint distributions. In this paper, a new method based on a binary-tree representation is introduced in order to store efficiently very large joint distributions. Our approach approximates any multi-dimensional and discrete-valued joint distribution using an adaptive discretization of the feature space defined by the random variables.

The discretization is based on the assumption that the regions in the space where the probability mass is high, have to be more accurately represented than the regions where the probability mass is low. The advantages of using such a tree representation are as following:

- the space complexity of the representation is reduced due to the preliminary assumption: less information is needed to store low mass probability regions, unless the probability tables which store an equal quantity of information for each region of the feature space.

- the number of points could grow dynamically, and thus, it is possible to refine the accuracy of the representation every time it is needed,

- every time a new point, which denotes an area of equal probability in the space, is inserted in the tree, the normalization constant is incrementally updated during the insertion of the point in the tree.

These advantages leads to two important facts:

- the probability of a random point drawn from the feature space is a sufficient statistic to represent the probability of the area around it. However, if another point is drawn near the first point, then the area is partitioned into smaller areas until the two points are separated into two distinct partitions.

- due to the incremental construction process of our approach, each time a new point is inserted in the tree, the probability distribution is refined. Such a representation is named an anytime representation: an approximate representation of the probability distribution is available each time it is needed. But, the accuracy of the represention grows along time [10, 15].

The method that will be presented in this paper has been successfully implemented in a larger system dedicated to inference in Bayesian networks and based on the Bayesian programming paradigm. This method is also a part of an European patent [3]. This system is largely used in our research team for Bayesian robots programming: approximate and efficient representation of large probability distributions is a success key for real-time robotics [9] since it allows to deal with complex Bayesian representation of the world and to fusion efficiently data issued from various sensors in order to control the robot. A more general survey of Bayesian programming is available in [1].

The organization of this paper is as follows. In Section 2, the multi-resolution binary trees (MRBT) are presented. After a description of general principles, an algorithm to construct such a tree is proposed. Examples of such trees and a discussion about time and space complexity will follow. Section 3 presents advantages in using MRBT for particular situations and discuss the current limitations of the model. The final section brings conclusion and opens up new perspectives.

## 2. Multi-resolution binary tree (MRBT)

### 2.1. Definition

A MRBT is a binary tree which partitions the feature space into a set of regions (or hypercube in the case of high-dimensional spaces) and fit a simple model in each one to represent the probability of this region. We choose a simple model to be more efficient: a point $X$ is drawn from the space and the probability of the region is: $P(R) = P(X).V_R$ where $V_R$ is the volume of the region. We made the assumption that the probability of the drawn point is a sufficient statistic to represent the probability of the region. Consequently, a region is represented by one and only one point. Besides, learning a MRBT is an incremental process where points are drawn sequentially from the feature space and used to characterizes small regions of the space. The more points there is, the smaller are the regions. The consequence of such an approach is that it is thus possible to refine incrementally the representation of the probability distribution.

Our approach is based on classical methods like *CART* (classification and regression trees) which makes a tree without any assumption on the size and form of the regions, thus leading to higher computational costs [4]. Other approaches to deal with probability distributions exist. For example, the probability tree [13] represents a probability distribution over a set of discrete random variables using a n-ary tree. Each inner node represents a variable of the joint distribution and each inner node has as many children as states the variable it represents has. For a given configuration of the variables, the real number stored in the leaf is the probability of the configuration if we reach the leaf by starting from the root node and for each inner node we follow the child corresponding to that configuration.

The accuracy of a probability tree is fixed by the total number of configurations of the set of variables. However, it is possible not to represent certain configurations of the variables leading to a smaller tree.

Then, a MRBT is a better tradeoffs between space and time because, it is possible to choose the accuracy of the representation. For a probability tree, each leaf is the probability of one and only one point (corresponding to a partic-ular configuration) in the feature space. For a MRBT, a leaf could contain either the probability of one point, either the probability of a region of the space.

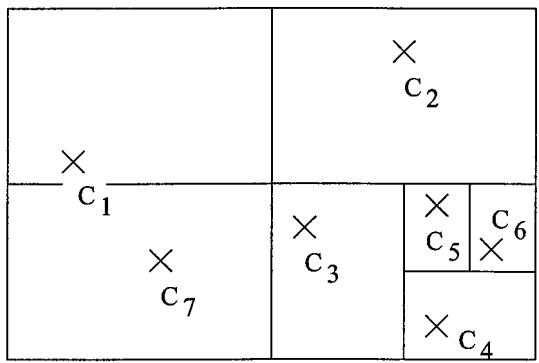

Figure 1. Dichotomic divided feature space

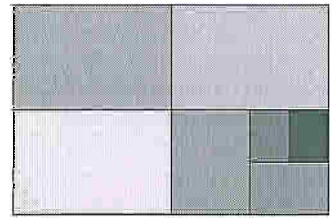

Figure 2. Approximative probability distribution based on the drawn points

## 2.2. A simple example

Let's consider a system described with two discrete random variables. The combined feature space of this two variables is a two-dimensional space, easily represented by a plan as shown in figure . Let's consider there exists a process which is able to randomly draw points from the space according to the probability distribution $P(X_1, X_2)$. A good drawing process would draw a sufficient quantity of points from areas where the probability mass is high and few points where it is low.

In order to approximate the probability distribution, the main assumption is that the probability of a drawn point is also the probability of the region it belongs to. A region is determined using a dichotomic process: for each point drawn from the space, the region it belongs to is divided into two smaller regions. The sub-region which contains the initial point holds its former probability, and the sub-region which contains the new point, holds the probability of the new point. By this way, the probability distribution is refined progressively.

Figure 2.2 shows the feature space of $X_1$ and $X_2$ divided by using several drawn points. Figure 2 shows the probability of each region. The gray level is a function of the probability of the region (the darker the region the higher the probability).

Moreover, the quality of the probability representation with a MRBT depends on the quality of the points drawing process. Popular methods are useful like an optimization method, an exhaustive covering of the feature space, a Monte-Carlo method, etc...

## 2.3. The learning process

The learning of a MRBT is an incremental process which has the anytime property: each time a new point is inserted, the MRBT holds an approximate representation of the probability distribution. This is useful when the learning time or the learning space are bounded. It is often the case in robotics where embedded computers have always serious limitations in term of computational and memory ressources. The learning process acts as follow:

1. the first drawn point don't partition the space and acts as the root of the tree;

2. a point $P$ is drawn from the feature space with a probability $p$;

3. if $P$ already exists then it is simply ignored;

4. if $P$ is a new point, then :

   (a) find the node (i.e. the region of the space) of the tree featuring a point $P_c$ and which contains the new point $P$,

   (b) split the region into two sub-regions, one dimension after the others. For example, if the space features three variables $X_1, X_2$ and $X_3$, then the space will be split depending on $X_1$, then $X_2$, then $X_3$, then $X_1$,etc...

   (c) the node (which was a leaf) becomes two new child nodes (which are new leafs), one featuring $P_C$ with a probability $p_c$ and the other featuring $P$ with a probability $p$.

   (d) If one leaf contains $P$ and $P_c$ again, then this leaf is recursively split into smaller regions according to the same process used for the MRBT construction and until $P$ and $P_c$ belong to two different region.

This process is repeated until no new points are available or the upper bound defined for the allocated memory or the time limit has been reached.

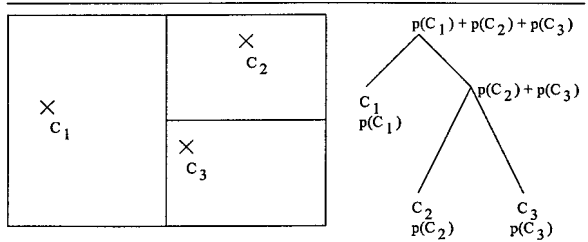

Figure 3. Simple example with only tree points

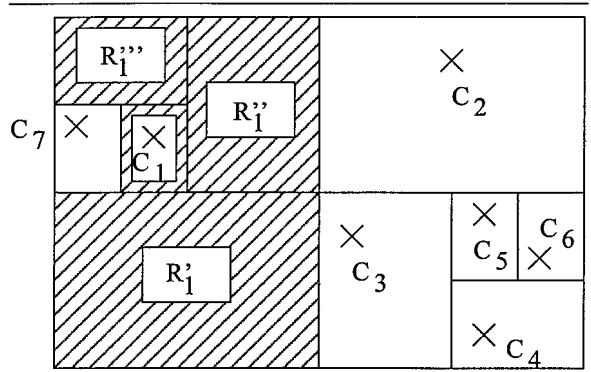

Figure 4. A more complex example with recursively learned regions

### 2.4. An more complex example

Let's consider the example in figure 3 (this is a reminder from figure 2.2). Each time a point is introduced in a region already containing an old point, this region is divided into two regions of equal dimensions. The split is made sequentially along each dimension, as shown in the previous section. The internal nodes become a likelihood equal to the sum of its children's probabilities. Accordingly, the root becomes the total likelihood of the tree which is used as the normalization constant. To reduce the computational cost during the learning phase, the probabilities associated to each node and leaf are never normalized.

Let's now consider the example in figure 4. The $C_7$ points is very near from $C_1$ and $C_7$ inducing several sub-regions with a probability proportional to the probability which contains $C_1$ at last. Solving such a conflict boils down to learn a MRBT where $C_1$ is the initial root of the MRBT. The region named $R'_1, R''_1, R'''_1$ have a probability equal to (resp.):

$$P(R'_1) = P(C_1)V_{R'_1}$$
$$P(R''_1) = P(C_1)V_{R''_1}$$
$$P(R'''_1) = P(C_1)V_{R'''_1}$$

The MRBT in figure 5 has two main partitions. The right subtree was generated with many real drawn points, $C_2, C_3, C_4, C_5, C_6$, while the left subtree was generated with only two points $C_1$ and $C_7$. This two points are very near and a lot of subdivisions are needed in order to accurately report the difference between $C_1$ and $C_7$ and because of the dichotomic approach.

## 3. Using MRBT

This new approach is very efficient in the domain of Bayesian reasoning and find many applications especially for reasoning under uncertainty. Indeed, it is possible to represent any discrete probability distribution without having to deal with huge probability tables. Efficiency is necessary when one has to compute and to use probability distributions in real time. In the case of probabilistic artificial intelligence, a probability distribution represent the knowledge on has about a fact or a conjunction of facts (joint probability distribution). Data fusion [2], artificial intelligence, robotics are all applications where it is necessary to deal with probabilistic and Bayesian reasoning. In general, basic uses of such distributions are essentially:

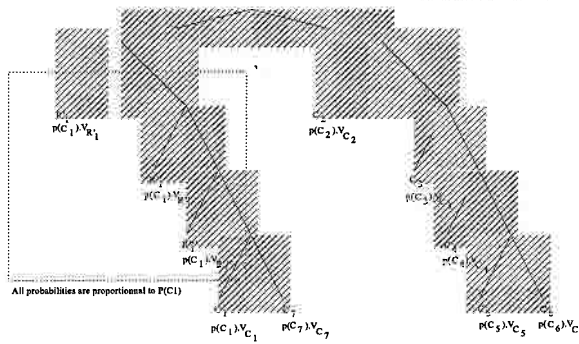

Figure 5. A MRBT with a conflicts between $C_1$ and $C_7$

- computing the probability of a particular point of the feature space,

- drawing points from the distribution,

- finding the best probability (for an action selection for example).

### 3.1. Probability of a particular point

In the case of a CPT, finding the probability $P(X)$ of a particular point $X$ is as simple as reading an array at a particular position. In the case of the MRBT, it is necessary to go through the binary tree. Given the coordinate of the point and given that each node of the tree contains characteristic information about the region it represents, it is easy to go all over the tree from the root node to a leaf. We have seen that the probability of each node is the probability of the region and is equal to:

$$P(R) = P(c_i).V_R$$

where $P(c_i)$ is the probability of the point which characterizes the region and $V_R$ is the volume of the region. When the leaf has been reached, the probability of the initial point $X$ is:

$$P(X) = \frac{P(R)}{P(root)}$$

where $P(root)$ is the normalization constant stored at the root of the tree. The main problem with this approach is that the complexity of finding a point is $O(\log(n))$ where $n$ is the number of drawn point used to make the MRBT. In the case of CPT, the same problem has trivially a complexity of $O(1)$. But the tradeoff between time and space using a MRBT allows to store complex probability distributions by using much less memory than CPTs.

### 3.2. Drawing a point

To draw a point from a MRBT, the algorithm is the same as finding the probability of a point. Let's consider $p_U$ a random value given by a uniform drawing process. Each level of the MRBT is a dichotomy on the repartition function of the probability distribution. At each node, the children which fits the best $p_U$ is chosen and the process follows on until a leaf is reached. Given the fact that each point of the leaf region has the same probability as each other point in this region, a point is chosen by uniformly and randomly drawing it from this region.

The complexity of this algorithm is $O(\log(n))$ which is better than using $F(X)$, the repartition function of the probability distribution $P(X)$ represented by the MRBT. Going all over a CPT to randomly draw a point according to a probability distribution has a complexity of $O(n)$.

### 3.3. Best probability

Finding the point with best probability is as easy as finding an occurrence of an element in any binary tree. But, during the construction of the MRBT, it is possible to store the leaf which currently corresponds to the maximum probability of the distribution. When one needs to draw a point of maximum probability, it is thus easy to draw a point from this previously stored region. If there is more than one region having the maximum probability, a linked list of maximum probability region is stored during the construction of the MRBT. Drawing a point of maximum probability is thus the same algorithm of that drawing a point in general. But in this case, the linked list is used in place of the whole MRBT.

## 4. Conclusion and future works

This paper has presented a novel approach to represent a probability distribution with many advantages:

- the storing space is much less high than the one needed when using probability tables,
- it is possible to refine the distribution by inserting more points as necessary,
- the space and time needed to construct a MRBT could be bounded in the case of real-time reasoning,
- it is possible to accurately approximate continuous probability distributions with any accuracy.

Based on classical methods like CART and probability trees, our approach adds the ability to have a useful approximate probability distribution at any time. The other main advantage is the possibility to refine the distribution every time it is needed.

Actually this representation is not still optimal in term of tradeoffs between time and space ressources. The main problem is related to the initial drawing process used to construct the MRBT. If the drawing process gives points in the same region, then the MRBT will be very unbalanced. If the drawing process gives points uniformly on the feature space, then the MRBT will be well-balanced but the accuracy will be not very good. Indeed, the regions with a low probability mass will be as accurately represented as those which have a high probability mass. Future works will be based on re-balancing and better sampling approach to ensure a well-balanced MRBT without a loss of accuracy in the probability distribution representation. This problem is also related to the ever growing size of the MRBT: it should be useful to decrease the size of the tree by pruning parts representing low probability mass region. This is necessary in memory-limited system (like a robot embedded computer) where the size of the tree is bounded by a hard limit. In this case, the problem is to prune parts of the tree which are not useful for the purpose of the system. Future works will focus on pruning and balancing strategies in order to decrease the time complexity for operations like finding the probability of a point and marginalizing a joint distribution.